\documentclass[letterpaper, 10 pt, conference]{ieeeconf}  

\IEEEoverridecommandlockouts                              

\usepackage{graphicx} 
\usepackage{epsfig} 
\usepackage{mathptmx} 
\usepackage{times} 
\usepackage{amsmath} 
\usepackage{amssymb}  
\usepackage{subcaption}
\usepackage{multirow}
\usepackage[normalem]{ulem}
\useunder{\uline}{\ul}{}
\usepackage{pifont}
\newcommand{\cmark}{\ding{51}}%
\newcommand{\xmark}{\ding{55}}%

\makeatletter
\let\NAT@parse\undefined
\makeatother
\usepackage{hyperref}
\hypersetup{
    colorlinks=true,
    linkcolor=blue,
    filecolor=magenta,      
    urlcolor=blue,
    }

\usepackage[capitalize]{cleveref}
\crefname{section}{Sec.}{Secs.}
\Crefname{section}{Section}{Sections}
\Crefname{table}{Table}{Tables}
\crefname{table}{Tab.}{Tabs.}

\usepackage{balance}

\title{\LARGE \bf
A Simple yet Effective Test-Time Adaptation for Zero-Shot Monocular Metric Depth Estimation
}

\author{
Rémi Marsal$^{1}$, Alexandre Chapoutot$^{1}$, Philippe Xu$^{1}$, David Filliat$^{1}$
\thanks{$^{1}$ Rémi Marsal, Alexandre Chapoutot, Philippe Xu and David Filliat are with \textit{U2IS, ENSTA Paris} 
\textit{Institut Polytechnique de Paris}
Palaiseau, France \newline
{\tt\small firstname.surname@ensta-paris.fr}}%
\thanks{This research was funded in whole or in part by the French National Research Agency (ANR) under the "ANR-23-MOXE-0003” project.}
}

\begin{document}

\maketitle
\thispagestyle{empty}
\pagestyle{empty}

\begin{abstract}

The recent development of \emph{foundation models} for monocular depth estimation such as Depth Anything paved the way to zero-shot monocular depth estimation. 
Since it returns an affine-invariant disparity map, the favored technique to recover the metric depth consists in fine-tuning the model.
However, this stage is not straightforward, it can be costly and time-consuming because of the training and the creation of the dataset.
The latter must contain images captured by the camera that will be used at test time and the corresponding ground truth.
Moreover, the fine-tuning may also degrade the generalizing capacity of the original model. 
Instead, we propose in this paper a new method to rescale Depth Anything predictions using 3D points provided by sensors or techniques such as low-resolution LiDAR or structure-from-motion with poses given by an IMU.
This approach avoids fine-tuning and preserves the generalizing power of the original depth estimation model while being robust to the noise of the sparse depth, of the camera-LiDAR calibration or of the depth model.
Our experiments highlight enhancements relative to zero-shot monocular metric depth estimation methods, competitive results compared to fine-tuned approaches and a better robustness than depth completion approaches.
Code available at \texttt{\href{https://github.com/ENSTA-U2IS-AI/depth-rescaling}{github.com/ENSTA-U2IS-AI/depth-rescaling}}.

\end{abstract}

\section{Introduction}

Despite the growth of 3D perception sensors such as LiDAR, time-of-flight, structured light or stereo cameras, for robotic systems, traditional monocular cameras remain a key sensor for any robot setup. In addition to being a cheaper solution, monocular depth estimation can also offer denser outputs as well as a larger depth range.
The increase in the number of open datasets for monocular depth estimation and the development of neural network architectures such as Vision Transformers that scale well with the size of the training dataset \cite{dosovitskiy2020image} allowed the emergence of foundation models for monocular depth estimation \cite{yang2024depth}.

Predicting metric depth (also referred to as \emph{absolute depth}) from a single image is fundamentally an impossible task due to scaling ambiguities.
However, monocular depth estimation methods \cite{bhat2021adabins} can achieve very good performances in a defined context.
These approaches have learned depth cues in images sampled from a certain distribution of environments that have been captured by a camera with fixed calibration.
Thus, such models, trained on a single dataset, generalize poorly to images taken with a different camera.
To avoid this issue during training on multiple datasets, some methods \cite{ranftl2020towards, yang2024depth, yang2024depthv2} learn an affine-invariant depth or disparity which allows impressive results on zero-shot relative depth estimation benchmarks.
Then, they propose to fine-tune their pre-trained models on the target domain to make metric depth predictions, \textit{i.e.}, on a dataset composed of images captured with the target camera calibration. 
Such a solution is costly in real cases due to the creation of the dataset and the training computation. Moreover, it must be performed for each new camera calibration.
Several solutions have been proposed to solve this issue by explicitly taking into account the camera calibration in the method \cite{guizilini2023towards} or trying to learn it \cite{piccinelli2024unidepth}. 
Depth completion methods propose another alternative \cite{zhang2023completionformer, tang2024bilateral} as they take as input some sparse depth measurements.
However, all these approaches may be more costly at inference or cannot be trained on image datasets with unknown calibration or unknown ground truth depth such as ImageNet \cite{deng2009imagenet} contrary to methods like Depth Anything \cite{yang2024depth, yang2024depthv2}.

In this paper, following depth completion approaches, we investigate a test-time adaptation that leverages sparse depth measurements for solving the scale ambiguity in order to perform zero-shot monocular metric depth estimation given affine-invariant disparity predictions.
Thus, an additional sensor is used to obtain some reference 3D points that are exploited to recover the scaling parameters.
For the sake of brevity, we will refer to \emph{rescaling} the process of finding an affine transformation and applying it to recover the metric depth. 
We focus our study on sparse depth that can be provided by other sources
including low-resolution LiDARs (with 16 and 32 beams), 2D LiDARs (with a single beam) that are often used for indoor robotics 
and structure-from-motion. In the latter, we assume a metric relative camera pose is given by an IMU. 
The advantage of our approach is twofold. 
On the one hand, it can be used with any monocular depth estimation model such as Depth Anything V1 and V2 \cite{yang2024depth, yang2024depthv2} with a high generalization ability due to their large and diverse training dataset. 
On the other hand, our method does not require any costly fine-tuning on the target domain and provides instant adaptation.
We conducted extensive experiments to evaluate our approaches on standard metric depth estimation benchmarks and demonstrate robustness to a noisy sparse depth, to errors in camera-LiDAR calibration or to a drop in sparse depth density.

\section{Related work}

\begin{figure*}[ht]
    \includegraphics[width=\textwidth]{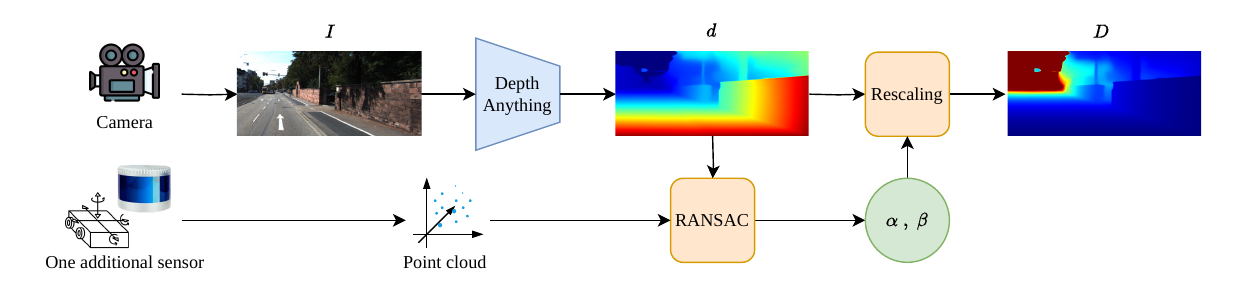}
    \caption{Illustration of our method. First, an affine-invariant disparity map $d$ is predicted from an image with a neural network such as Depth Anything \cite{yang2024depth} while in parallel a sensor is used to estimate a set of 3D points $P$. $P$ and the corresponding values in $d$ are then used to estimate the scaling parameters $\alpha$ and $\beta$ using a RANSAC \cite{fischler1981random}. The parameters are applied to $d$ to recover the metric depth $D$.}
    \label{fig:diagram}
\end{figure*}

\subsection{Monocular depth estimation}

While pioneer works on monocular depth estimation relied on Markov Random Fields \cite{saxena2005learning}, subsequent ones have shown the effectiveness of convolutional neural networks~\cite{eigen2014depth, laina2016deeper} then transformers for this task \cite{ranftl2021vision}. More recently, benefiting from advances in image generation \cite{rombach2022high}, impressive results have been obtained using diffusion models \cite{saxena2024surprising, ke2024repurposing}.
Regarding the output, monocular depth estimation has been initially addressed as a regression problem \cite{eigen2014depth}, before moving to classification approaches with discrete bins that show better performances \cite{bhat2021adabins, bhat2022localbins}.
Monocular depth estimation methods can also be divided into whether they are supervised \cite{eigen2014depth, laina2016deeper, bhat2021adabins} or self-supervised \cite{godard2019digging}. 
Recently, the multiplication of depth estimation benchmark allowed training models on multiple domains at once \cite{ranftl2020towards} and paved the way to zero-shot monocular depth estimation. Due to the inherent scale ambiguity of depth estimation from a single image, these methods are mostly trained to produce affine-invariant depth or disparity predictions \cite{ranftl2020towards, ranftl2021vision, yang2024depth}.
Our work aims to be used with any depth estimation model as long as it returns disparity maps that are accurate within an affine transformation whatever its other characteristics in terms of training set or architecture.

\subsection{Scale estimation for monocular depth}

A major issue of monocular depth prediction relies on scale ambiguity which means that the true size of an object cannot be recovered for sure from a single image.
Since most monocular depth estimation models are trained for a specific camera calibration, using them with another camera leads to ill-scaled predictions.
The most common way to recover the metric depth in this situation is to fine-tune the model on a dataset collected with the camera that will be used at inference \cite{ranftl2020towards, ranftl2021vision}. In practice, this solution is costly to implement as it requires the creation of an image dataset with the relative ground truth and a new training.
Other works focus on the temporal consistency of the scale of depth predictions \cite{zhang2019exploiting, li2021enforcing}. 
Closer to our work, \cite{wofk2023monocular} learns to predict scaled depth maps from affine-invariant disparity maps and visual-inertial odometry. 
Also, there exists an extensive literature on depth completion which studies neural network architectures that take as inputs both an image and a sparse depth map \cite{ma2018sparse, zhang2023completionformer, tang2024bilateral}.
In \cite{dana2024more}, authors propose to estimate the scale factor for a target domain from a model that has been jointly trained 
on a source domain with known ground truth and on a target domain without depth annotation.
In contrast, we propose to rescale at test time any disparity prediction that is correct up to an affine transformation with no additional training or fine-tuning but by leveraging reference 3D points provided by an external sensor or technique.
Furthermore, relying on external sensor makes our approach adaptable to any camera calibration.

\subsection{Zero-shot monocular metric depth estimation}

ZoeDepth \cite{bhat2023zoedepth} is the first zero-shot monocular metric depth estimation method. It first consists in relative depth pre-training of the MiDaS \cite{ranftl2020towards} backbone then fine-tuning two metric bin modules, one for indoor scenes and the other for outdoor scenes. 
More recently, other methods have been proposed without any fine-tuning.
Thus, ScaleDepth \cite{zhu2024scaledepth} decomposes metric depth estimation in relative depth estimation and scale estimation each with a dedicated module. It can also leverage a text description of the scene to guide the supervision.  
In \cite{guizilini2023towards}, authors introduce a dedicated architecture that takes as input the calibration matrix in addition to the image.
On the contrary, \cite{yin2023metric3d, hu2024metric3d} make predictions from the images only but applies transformation on the input images so the predictions are invariant to the image size or the camera calibration.
UniDepth \cite{piccinelli2024unidepth} estimates an internal representation of the camera calibration directly from the input images only.
These approaches still have drawbacks since they are often more costly at inference or need the calibration of the images even at the training stage. This prevents exploiting image datasets with unknown calibration unlike Depth Anything V1 \cite{yang2024depth} and V2 \cite{yang2024depthv2} which leveraged a distillation strategy with such datasets to improve their generalization abilities.

\section{Method}

Let $\phi$ be a monocular depth estimation model such as \cite{ranftl2020towards, yang2024depth} that is trained to predict an affine-invariant disparity map $d \in \mathbb{R}^{H \times W}$ given an input
RGB image $I \in \mathbb{R}^{H \times W \times 3}$ where $H$ and $W$ are the height and the width of the image $I$. Therefore, the metric or absolute depth map $D \in \mathbb{R}^{H \times W}$ that corresponds to the inverse of the metric or absolute disparity $D^{-1}$ is given by the relation:
    \begin{equation}
        D^{-1} = \alpha d + \beta,
    \end{equation}
where parameters $\alpha\in\mathbb{R}^+_*$ and $\beta\in\mathbb{R}$ are the unknown scaling factor and the offset, respectively.

Our method, illustrated in \cref{fig:diagram}, aims at recovering the metric depth map $D$ at test time from the affine-invariant disparity map $d$ and a set of $N$ reference 3D points $P \in \mathbb{R}^{N \times 3}$ by regressing the parameters $\alpha$ and $\beta$.
First, we perform a bilinear sampling on the affine-invariant map $d$ at the locations of the projection of the reference 3D points $P$ on the image plan so as to have an affine-invariant disparity value for each reference 3D point. 
Second, we leverage linear regression to estimate $\alpha$ and $\beta$ parameters.
To favor the robustness against potential outliers in the reference 3D points, we use a RANSAC algorithm \cite{fischler1981random}, but other robust regression methods could be considered.

In our approach, we assume the set $P$ is provided by a low-resolution LiDAR or a structure-from-motion (SFM) technique in which poses are given by an IMU.
When SFM is leveraged, having metric poses is necessary to triangulate the absolute coordinates of matching points in consecutive video images.
Thus, $P$ may be a very sparse depth map, \textit{i.e.}, $N \ll HW$ and may also contain some outlier measurements such as those caused by LiDAR reflections from dust or artifacts introduced by SFM in dynamic scenes.

As our method relies on affine-invariant disparity maps, we can exploit, without any fine-tuning, \emph{foundation models} like Depth Anything V1 or V2 \cite{yang2024depth,yang2024depthv2} that are trained to predict such outputs.
Furthermore, the robustness to outliers enabled by the RANSAC allows low-quality sensors to be used to obtain the reference 3D points. Thus, we can provide metric depth maps whatever the environment or the camera with no fine-tuning.
Nevertheless, since Depth Anything processes each image independently and is supervised in such a way there is no guarantee its disparity predictions are normalized identically even for consecutive images of a video, the scaling factor $\alpha$ and the offset $\beta$ must be estimated for each image.

\section{Experiments}

\subsection{Dataset and metrics}

To evaluate performance in zero-shot monocular metric depth estimation, we focus on standard monocular depth estimation benchmarks that have not been used to pretrain Depth Anything \cite{yang2024depth}. This includes indoor datasets: NYUv2~\cite{nyu}, SUN-RGBD~\cite{song2015sun}, IBIMS-1~\cite{ibims} and DIODE indoor \cite{diode_dataset} and outdoor datasets: KITTI \cite{kitti} , DDAD \cite{guizilini20203d} and DIODE outdoor \cite{diode_dataset}. 
We adopt standard depth estimation metrics (see \cite{eigen2014depth, yang2024depth}) to compare our method to other approaches of the literature:

\begin{equation}
    \text{RMS} = \sqrt{\frac{1}{|\Omega|}\sum_{p \in \Omega}(\hat{D}(p) - D^*(p) )^2}
\end{equation}

\begin{equation}
    \text{AbsRel} = \frac{1}{|\Omega|}\sum_{p \in \Omega}\frac{|\hat{D}(p) - D^*(p)|}{D^*(p)}
\end{equation}

\begin{equation}
    \delta_1 = \frac{1}{|\Omega|} \Bigg|  \Bigg\{ p \in \Omega \bigg| \frac{1}{1.25} < \frac{\hat{D}(p)}{D^*(p)} < \frac{1.25}{1} \Bigg\} \Bigg|
\end{equation}

\noindent where $\Omega$ is the set of pixels for which the ground truth is available and $|.|$ applied to a set returns its cardinal and the absolute value of a scalar otherwise.
For a pixel $p \in \Omega$, $\hat{D}(p)$ is the estimation corresponding the ground truth depth $D^*(p)$.

\subsection{Implementation details}

All experiments have been conducted using Depth Anything V1 \cite{yang2024depth} with ViT Large \cite{dosovitskiy2020image} without fine-tuning unless otherwise mentioned. 
We adopt Depth Anything V1 \cite{yang2024depth} code base and settings for evaluation except for comparison with depth completion methods as detailed later.
We simulate low-resolution LiDARs by evenly selecting as many horizontal lines in the ground truth depth maps as the number of beams in 32-laser, 16-laser or 2D LiDARs (\textit{i.e.}, with a single beam).
In contrast, no ground truth depth is used as input when rescaling with SFM.
We study rescaling with SFM for KITTI~\cite{kitti} and DDAD datasets \cite{guizilini20203d} only as they both consist of temporal sequences of images including the pose estimations. 
To obtain the reference 3D points, we first extract matching keypoints in the target image and the previous one using SIFT \cite{lowe1999object} or OmniGlue \cite{jiang2024omniglue} and triangulate them using the pose between these images.
Since this strategy requires enough displacement magnitude between the two images, we only consider image couples with a rotation higher than $5$ degrees or a translation greater than $1.5$ and $2$ meters for KITTI and DDAD datasets, respectively. We note that this threshold is the only hyperparameter that needs to be tuned in our approach.

\subsection{Comparison with monocular depth esimation methods}

\begin{table}[ht]
    \centering
    \caption{Quantitative results on the NYUv2 dataset \cite{nyu} (indoor). (ZS) means Zeros-Shot, (FT) stands for Fine-Tuned on NYUv2.
    }
    \begin{tabular}{|c|ccc|}
        \hline
        Methods                                       & $\delta_1$ $\uparrow$ & AbsRel $\downarrow$ & RMS $\downarrow$ \\ \hline
        ZeroDepth \cite{guizilini2023towards} (ZS)    & 0.926          & 0.081          & 0.338          \\
        Metric3D \cite{yin2023metric3d} (ZS)          & 0.944          & 0.083          & 0.310          \\
        Metric3D V2 \cite{hu2024metric3d} (ZS)        & 0.975          & 0.063          & 0.251          \\
        Unidepth \cite{piccinelli2024unidepth} (ZS)   & {\ul 0.984}    & 0.058          & {\ul 0.201}    \\ \hline
        ZeroDepth \cite{guizilini2023towards} (FT)    & 0.954          & 0.074          & 0.269          \\
        ZoeDepth \cite{bhat2023zoedepth} (FT)         & 0.955          & 0.075          & 0.270          \\
        Metric3D V2 \cite{hu2024metric3d} (FT)        & \textbf{0.989} & 0.047          & \textbf{0.183}  \\
        ScaleDepth \cite{zhu2024scaledepth} (FT)      & 0,957          & 0,074          & 0,267          \\
        Depth Anything \cite{yang2024depth} (FT)      & {\ul 0.984}    & 0.056          & 0.206          \\
        Depth Anything V2 \cite{yang2024depthv2} (FT) & {\ul 0.984}    & 0.056          & 0.206          \\ \hline
        Ours w/ LiDAR 1 beam                          & 0.939          & 0.063          & 0.652          \\
        Ours w/ LiDAR 16 beams                        & 0.976          & \textbf{0.039} & 0.454             \\
        Ours w/ LiDAR 32 beams                        & 0.974          & {\ul 0.040}    & 0.461          \\ \hline
        \end{tabular}%
    \label{tab:nyu}
\end{table}

\begin{table}[ht]
    \centering
        \caption{Quantitative results on the KITTI dataset \cite{kitti} (outdoor). 
    (ZS) means Zeros-Shot, (FT) stands for Fine-Tuned on KITTI.}
    \begin{tabular}{|c|ccc|}
    \hline
    Methods                                              & $\delta_1$ $\uparrow$ & AbsRel $\downarrow$ & RMS $\downarrow$ \\ \hline
    ZeroDepth \cite{guizilini2023towards} (ZS)           & 0.910          & 0.102          & 4.044 \\
    Metric3D \cite{yin2023metric3d} (ZS)                 & 0.964          & 0.058          & 2.770 \\
    Metric3D V2 \cite{hu2024metric3d} (ZS)               & 0.974          & 0.052          & 2.511 \\ \hline
    ZoeDepth \cite{bhat2023zoedepth} (FT)                & 0.971          & 0.057          & 2.281 \\
    ZeroDepth \cite{guizilini2023towards} (FT)           & 0.968          & 0.053          & 2.087 \\
    Metric3D V2 \cite{hu2024metric3d} (FT)               & \textbf{0.985} & \textbf{0.044} & 1.985 \\
    ScaleDepth \cite{zhu2024scaledepth} (FT)             & 0,980          & 0,048          & 1,987 \\
    Depth Anything \cite{yang2024depth} (FT)             & 0.982          & 0.046          & {\ul 1.869} \\
    Depth Anything V2 \cite{yang2024depthv2} (FT)        & {\ul 0.983}    & {\ul 0.045}    & \textbf{1.861} \\ \hline
    Ours w/ LiDAR 1 beam                                 & 0.891          & 0.131          & 3.096 \\
    Ours w/ LiDAR 16 beams                               & 0.967          & 0.060          & 2.695 \\
    Ours w/ LiDAR 32 beams                               & 0.967          & 0.060          & 2.673 \\
    Ours w/ SFM (SIFT \cite{lowe1999object})             & 0.893          & 0.103          & 3.920 \\
    Ours w/ SFM (OmniGlue \cite{jiang2024omniglue})      & 0.925          & 0.093          & 3.562 \\
    \hline
    \end{tabular}%
    \label{tab:kitti}
\end{table}

\begin{table*}[ht]
    \centering
        \caption{Quantitative results on different zero-shot indoor benchmarks. FT NYUv2 stands for Fine-Tuned on NYUv2~\cite{nyu}.}
    \begin{tabular}{|c|ccc|ccc|ccc|}
        \hline
        \multirow{2}{*}{Methods}                                & \multicolumn{3}{c|}{SUN-RGBD \cite{song2015sun}}               & \multicolumn{3}{c|}{IBIMS-1 \cite{ibims}}      & \multicolumn{3}{c|}{DIODE Indoor \cite{diode_dataset}} \\
         &
          $\delta_1$ $\uparrow$ &
          AbsRel $\downarrow$ &
          RMS $\downarrow$ &
          $\delta_1$ $\uparrow$ &
          AbsRel $\downarrow$ &
          RMS $\downarrow$ &
          $\delta_1$ $\uparrow$ &
          AbsRel $\downarrow$ &
          RMS $\downarrow$ \\ \hline
        Metric3D \cite{yin2023metric3d}     & --          & --             & --          & --          & 0.144       & --    & --        & 0.252     & --        \\
        Metric3D V2 \cite{hu2024metric3d}                      & --          & --             & --          & --          & 0.185       & 0.592 & --        & 0.093     & 0.389     \\
        Unidepth \cite{piccinelli2024unidepth}                         & \textbf{0.966}       & --             & --          & 0.797       & --          & --    & 0.774     & --        & --        \\
        ZoeDepth \cite{bhat2023zoedepth} (FT NYUv2)             & 0.864       & \textbf{0.119} & {\ul 0.346} & 0.658       & 0.169       & 0.711 & 0.4       & 0.324     & 1.581     \\
        ScaleDepth \cite{zhu2024scaledepth} (FT NYUv2)           & 0.864       & {\ul 0.127}    & 0.360       & 0.788       & 0.156       & 0.601 & 0.455     & 0.277     & 1.350      \\
        Depth Anything \cite{yang2024depth} (FT NYUv2)                & 0.658       & 0.500          & 0.616       & 0.714       & 0.150       & 0.593 & 0.303     & 0.325     & 1.476     \\ \hline
        Ours w/ LiDAR 1 beam & 0.924 & 0.281          & 0.357       & {\ul 0.942} & {\ul 0.072} & 0.340 & 0.934     & 0.098     & 0.411     \\
        Ours w/ LiDAR 16 beams &
          {\ul 0.951} &
          0.275 &
          \textbf{0.295} &
          \textbf{0.979} &
          \textbf{0.037} &
          {\ul 0.232} &
          \textbf{0.953} &
          {\ul 0.084} &
          {\ul 0.361} \\
        Ours w/ LiDAR 32 beams &
          {\ul 0.951} &
          0.279 &
          \textbf{0.295} &
          \textbf{0.979} &
          \textbf{0.037} &
          \textbf{0.231} &
          {\ul 0.952} &
          \textbf{0.083} &
          \textbf{0.359} \\ \hline
        \end{tabular}%
    \label{tab:zs-indoor}
\end{table*}

\begin{table*}[ht]
\centering
\caption{Quantitative results on different zero-shot outdoor benchmarks. FT KITTI stands for Fine-Tuned on KITTI \cite{kitti}.}
\label{tab:zs-outdoor}
\begin{tabular}{|c|ccc|ccc|}
\hline
\multirow{2}{*}{Methods}                                          & \multicolumn{3}{c|}{DIODE Outdoor \cite{diode_dataset}}               & \multicolumn{3}{c|}{DDAD \cite{guizilini20203d}}                        \\
 &
  $\delta_1$ $\uparrow$ &
  AbsRel $\downarrow$ &
  RMS $\downarrow$ &
  $\delta_1$ $\uparrow$ &
  AbsRel $\downarrow$ &
  RMS $\downarrow$ \\ \hline
ZoeDepth (FT KITTI) \cite{bhat2023zoedepth}                      & --             & --             & --             & 0.835          & 0.129          & 7.108         \\
ZeroDepth \cite{guizilini2023towards}                            & --             & --             & --             & 0.814          & 0.156          & 10.678         \\
Metric3D \cite{yin2023metric3d}                                                        & --             & {\ul 0.414}    & 6.934          & --             & --             & --             \\
Metric3D V2 \cite{hu2024metric3d}                                & --             & \textbf{0.221} & \textbf{3.897} & --             & --             & --             \\
Unidepth \cite{piccinelli2024unidepth}                           & --             & --             & --             & 0.864          & --             & --             \\
ScaleDepth \cite{zhu2024scaledepth} (FT KITTI)                   & 0.333          & 0.605          & 6.950          & 0.863          & 0.120          & 6.378          \\
Depth Anything V1 \cite{yang2024depth} (FT KITTI)                            & 0.288          & 0.794          & 6.641          & 0.886          & 0.105          & 5.931          \\ \hline
Ours w/ LiDAR 1 beam                                      & 0.689          & 0.880          & 6.222          & 0.706          & 0.326          & 9.229          \\
Ours w/ LiDAR 16 beams                                    & {\ul 0.796}    & 0.697          & 4.933          & {\ul 0.897}    & {\ul 0.097}    &  {\ul 3.716}          \\
Ours w/ LiDAR 32 beams                                    & \textbf{0.799} & 0.683          &  {\ul 4.835}          & {\ul 0.897}    & \textbf{0.096} & \textbf{3.675}          \\
Ours w/ SFM (SIFT)                      & --             & --             & --             & 0.776          & 0.161          & 5.929          \\
Ours w/ SFM (Omniglue)                  & --             & --             & --             & \textbf{0.947} & 0.112          & 5.300          \\ \hline
\end{tabular}%
\end{table*}

In \cref{tab:nyu}, we compare our LiDAR-based rescaling approach with zero-shot monocular depth estimation methods and other depth estimation methods that have been fine-tuned on the NYUv2 dataset \cite{nyu}.
We conducted a similar study in \cref{tab:kitti} on the KITTI dataset \cite{kitti} with in addition results of rescaling with 3D reference points provided by structure-from-motion. 
\cref{tab:zs-indoor} and \cref{tab:zs-outdoor} provide zero-shot performance on indoor and outdoor datasets, respectively. For each dataset, our results show rescaling using LiDARs with different numbers of beams. 
Additionally, we present results of rescaling with structure-from-motion for the DDAD dataset \cite{guizilini20203d}.
We do not provide results of rescaling with structure-from-motion on NYUv2~\cite{nyu}, SUN-RGBD~\cite{song2015sun}, IBIMS-1~\cite{ibims} and DIODE Indoor and Outdoor \cite{diode_dataset} because they do not consist of temporal image sequences with poses.
The results of the monocular depth estimation methods we compare against were directly taken from their respective papers, which explain missing numbers when they have not been reported by their authors.

\paragraph{Rescaling with LiDAR} Our experiments highlight an overall benefit of using our rescaling approach with $32$-laser or $16$-laser LiDAR rather than using zero-shot monocular metric depth estimation. Thus, we observe on average $46$\% improvement on $\delta_1$, $0.5$\% on AbsRel and $47$\% on RMS relative to the other zero-shot methods.
In contrast, the methods that have been fine-tuned on the same domain as the test set (see lines with \emph{(FT)} in \cref{tab:nyu} and \cref{tab:kitti}) compare favorably to ours. These approaches benefit from an additional in-domain training which is often costly due to dataset creation and training computation.  
We notice that our method is sensitive to the domain with on average $28$\% and $13$\% enhancements on indoor (including NYUv2) and outdoor (including KITTI) datasets, respectively, relative to other zero-shot methods whatever the metric. 
Regarding performance with 2D LiDAR, the same sensitivity to the domain is apparent.
However, the results are a bit lower than zero-shot and fine-tuned metric depth estimation methods and lower than our other approaches, especially on outdoor datasets while remaining competitive on indoor datasets.
Interestingly, we observe that increasing the number of beams from 16 to 32 does not necessarily improve the performance.

\paragraph{Rescaling with structure-from-motion} Our experiments highlight that structure-from-motion techniques can provide reliable reference points for rescaling Depth Anything affine-invariant disparity predictions. 
More specifically, we show that results with SIFT \cite{lowe1999object} are slightly lower than the ones of other zero-shot monocular metric depth estimation methods, while using OmniGlue \cite{jiang2024omniglue} increases performance relative to other zero-shot methods of the $\delta_1$, AbsRel and RMS metrics on average by $7$\%, $4$\% and $19$\%, respectively. 
Compared to the other approaches, rescaling with SFM performs better than rescaling with LiDAR 1 beam but worse than other methods that may be more complex to set up including rescaling with LiDAR 16 and 32 beams or fine-tuned approaches.
However, such approaches require LiDAR camera calibration or an additional training on a dataset that needs be created in real cases.

\subsection{Comparison with depth completion methods}

\begin{table*}[ht]
\centering
\caption{Quantitative comparison with depth completion methods on different zero-shot outdoor benchmarks. }
\label{tab:dc-zs-outdoor}
\begin{tabular}{|c|ccc|ccc|}
\hline
\multirow{2}{*}{Methods}                        & \multicolumn{3}{c|}{DIODE Outdoor \cite{diode_dataset}}      & \multicolumn{3}{c|}{DDAD \cite{guizilini20203d}}                        \\
 &
  $\delta_1$ $\uparrow$ &
  AbsRel $\downarrow$ &
  RMS $\downarrow$ &
  $\delta_1$ $\uparrow$ &
  AbsRel $\downarrow$ &
  RMS $\downarrow$ \\ \hline
CompletionFormer \cite{zhang2023completionformer} w/ LiDAR 64 beams &
  0.679 &
  \textbf{0.439} &
  6.323 &
  0.711 &
  0.173 &
  8.612 \\
BP-Net \cite{tang2024bilateral} w/ LiDAR 64 beams & --             & --    & --             & {\ul 0.881}    & \textbf{0.075} & \textbf{0.825} \\
Ours w/ LiDAR 64 beams                  & \textbf{0.800} & 0.677 & \textbf{4.870} & \textbf{0.933} & {\ul 0.078}    & {\ul 5.283}    \\ \hline
\end{tabular}%
\end{table*}

\begin{table*}[ht]
\centering
\caption{Quantitative comparison with depth completion methods on different zero-shot indoor benchmarks. }
\label{tab:dc-zs-indoor}
\begin{tabular}{|c|ccc|ccc|ccc|}
\hline
\multirow{2}{*}{Methods}         & \multicolumn{3}{c|}{SUN-RGBD \cite{song2015sun}} & \multicolumn{3}{c|}{IBIMS-1 \cite{ibims}}               & \multicolumn{3}{c|}{DIODE Indoor \cite{diode_dataset}}       \\
 &
  $\delta_1$ $\uparrow$ &
  AbsRel $\downarrow$ &
  RMS $\downarrow$ &
  $\delta_1$ $\uparrow$ &
  AbsRel $\downarrow$ &
  RMS $\downarrow$ &
  $\delta_1$ $\uparrow$ &
  AbsRel $\downarrow$ &
  RMS $\downarrow$ \\ \hline
CompletionFormer \cite{zhang2023completionformer} w/ 500 random samples &
  \textbf{0.964} &
  \textbf{0.129} &
  \textbf{0.243} &
  \textbf{0.967} &
  \textbf{0.080} &
  0.315 &
  \textbf{0.958} &
  0.252 &
  0.403 \\
BP-Net \cite{tang2024bilateral} w/ 500 random samples & --       & --       & --      & 0.953       & {\ul 0.093} & \textbf{0.039} & --    & --             & --             \\
Ours w/ 500 random samples & 0.936    & 0.288    & 0.324   & {\ul 0.951} & 0.101       & \textbf{0.039} & 0.941 & \textbf{0.092} & \textbf{0.402} \\ \hline
\end{tabular}%
\end{table*}

\begin{figure*}
    \centering
    \begin{subfigure}{\textwidth}
        \includegraphics[width=\textwidth]{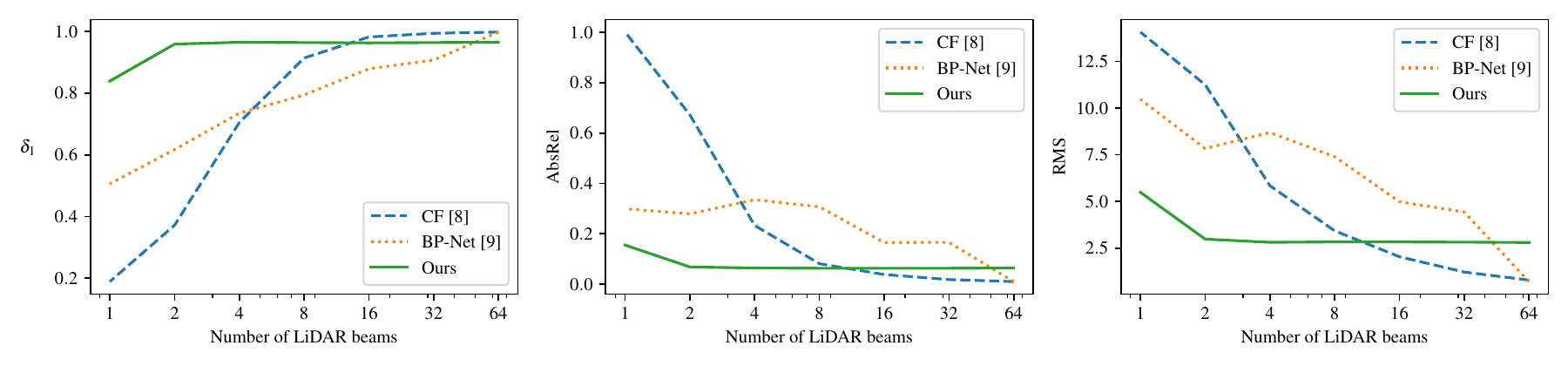}
    \end{subfigure}
    
    \begin{subfigure}{\textwidth}
        \includegraphics[width=\textwidth]{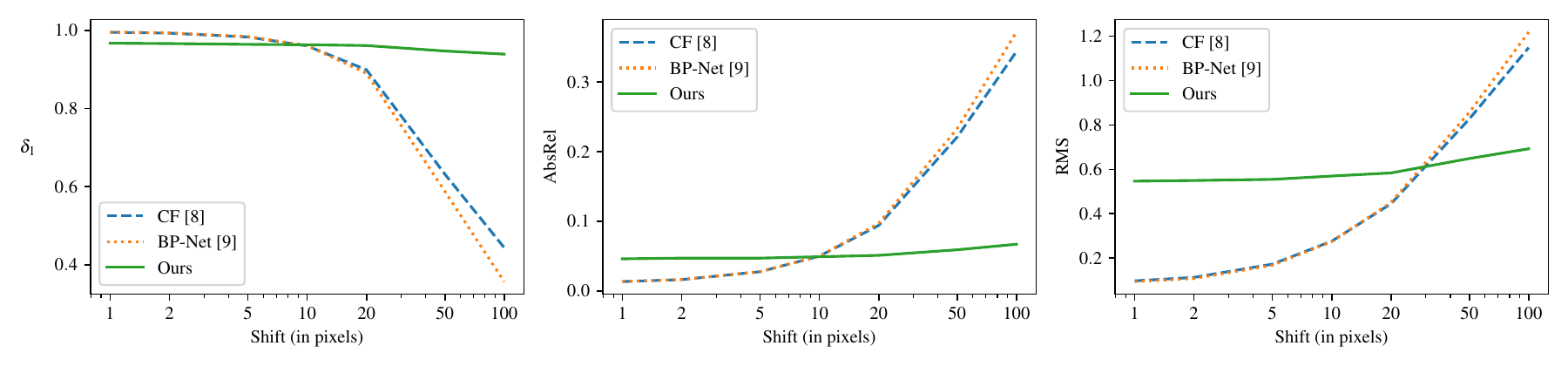}
    \end{subfigure}
    
    \begin{subfigure}{\textwidth}
        \includegraphics[width=\textwidth]{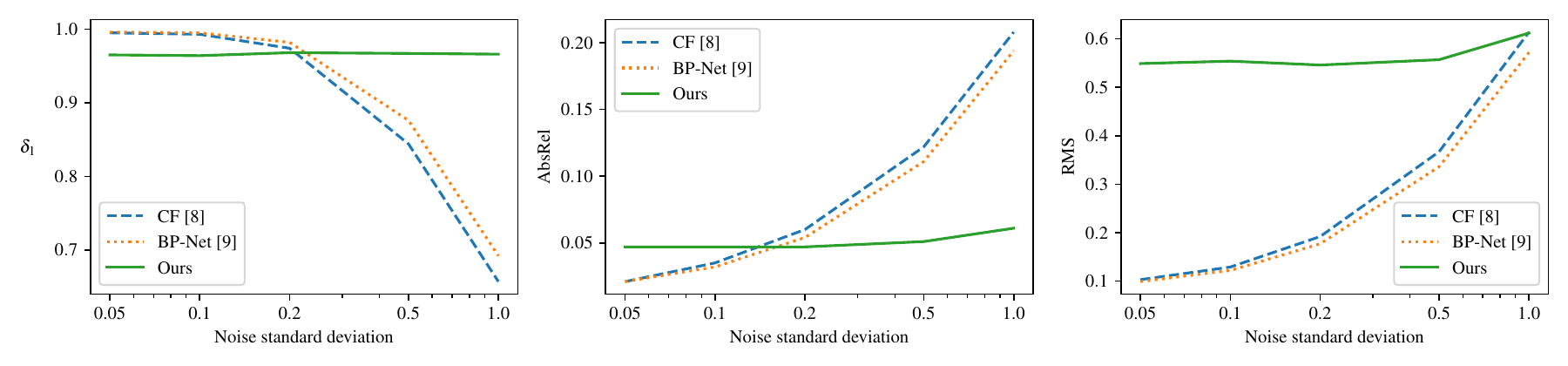}
    \end{subfigure}
    
    \caption{Quantitative study of the robustness of depth completion methods: (top) impact of the number of LiDAR beams on performance on KITTI \cite{kitti}, (middle) impact of camera-LiDAR calibration error by applying a horizontal or vertical shift to depth samples on NYUv2 \cite{nyu}, (bottom) impact of random noise on depth samples on the performance on NYUv2 \cite{nyu}.}
    \label{fig:dc}
\end{figure*}

In this section, we compare our method with two recent depth completion approaches, CompletionFormer \cite{zhang2023completionformer} and BP-Net \cite{tang2024bilateral}. 
Both methods have been trained once on KITTI~\cite{kitti} and once on NYUv2~\cite{nyu}.
For the sake of fairness, we conducted evaluations with the settings used to train CompletionFormer \cite{zhang2023completionformer} and BP-Net \cite{tang2024bilateral}.
Thus, for NYUv2 and other indoor benchmarks, images are resized to $320 \times 240$ and then center-cropped to $304 \times 228$, the sparse depth is obtained by randomly sampling 500 points in the ground truth depth.
For KITTI and other outdoor benchmarks, images are cropped to $1216 \times 256$ and the number of LiDAR beams is set to 64. 
We note that no evaluation of BP-Net \cite{tang2024bilateral} is performed on SUN-RGBD \cite{song2015sun}, DIODE Indoor and DIODE Outdoor \cite{diode_dataset} since the camera calibration of the input images that is required by BP-Net is absent in those datasets.

\paragraph{Zero-shot depth completion} We evaluate zero-shot performance on outdoor and indoor benchmarks in \cref{tab:dc-zs-outdoor} and \cref{tab:dc-zs-indoor} with the depth completion networks trained on KITTI \cite{kitti} and NYUv2 \cite{nyu}, respectively. 
For outdoor datasets, the results show that our method performs on par with CompletionFormer \cite{zhang2023completionformer} on DIODE Outdoor \cite{diode_dataset} and is better on DDAD \cite{guizilini20203d} for all metrics while BP-Net~\cite{tang2024bilateral} appears to be advantageous on the latter benchmark.
Regarding indoor datasets, performance on IBIMS-1 \cite{ibims} and DIODE indoor are close. The fact that CompletionFormer outperforms our method on SUN-RGBD may be partially explained by very similar domains with NYUv2 \cite{nyu} (used to train CompletionFormer) since one of the cameras used in SUN-RGBD is the same as that of NYUv2.

\paragraph{Robustness to reduction of the sparse depth density} The cases previously analyzed represent ideal conditions, as the test settings precisely match those of the training stage. However, in real cases, test conditions may be imposed or may change. 
If they differ significantly from those used during the training of any available depth completion model, retraining may be necessary to ensure reliable performance.
We illustrate this point by examining the robustness of depth completion methods to changes in the distribution of the sparse depth. Indeed, in practical scenarios, the sparse depth distribution is given by the type of LiDAR used. Additionally, when SFM is employed, variations in image texture can cause fluctuations in sparse depth density over time.
\cref{fig:dc} (top) show the influence of decreasing the number of LiDAR beams (from 64 to 1) on the KITTI Depth Completion dataset~\cite{kitti_dc}.
The results highlight the robustness of our method to the reduction of the sparse depth density relative to depth completion techniques, as their performance drops when ours is barely affected.
The advantage of our method comes from its simplicity since only two 3D points are necessary in theory to regress the two rescaling parameters.
However, depth completion robustness is likely to be improved by randomly varying the density of the sparse depth at training.
We note that the better results of depth completion methods on a large number of samples are likely to be due to the evaluation sets which come from the same dataset as their training data contrary to our approach.

\paragraph{Robustness to camera-LiDAR calibration errors} A major weakness of our approach and of depth completion methods is their reliance on camera-LiDAR calibration. In \cref{fig:dc} (middle), we compare the sensitivity of our method to camera-LiDAR calibration errors relative to depth completion approaches. We simulate these errors by shifting the sparse depth relative to the image in a random direction (left, right, up or down) by a defined number of pixels. We choose this setup for its ease of implementation though a more realistic setup could be considered. We find that depth completion approaches perform significantly worse with misalignment between the sparse depth and the image, whereas our method is resilient to camera-LiDAR calibration errors.
However, appropriate data augmentation could improve the performance of depth completion methods.
The good results of depth completion methods in the absence of calibration errors is likely due to overfitting on the NYUv2 dataset.

\paragraph{Robustness to noisy sparse depth} Another limitation of our analysis on zero-shot depth completion is that the sparse depth corresponds to the ground truth while in real cases, it may be noisy.
To assess the robustness of depth completion methods to noise, we conduct experiments on the NYUv2 dataset \cite{nyu} where a centered Gaussian noise is added to the inputs sparse depth. We study the impact of gradually increasing its standard deviation from 5cm to 1 meter. 
We choose Gaussian noise for its simplicity but other types of noise commonly encountered in SFM or LiDAR could have been considered. 
The results in \cref{fig:dc} (bottom) demonstrate the robustness of our approach relative to noisy depth samples.
Unlike other depth completion approaches, which suffer from significant performance degradation as noise increases, our method remains stable.
However, the robustness to noisy sparse depth could be improved by adding random noise at training stage as data augmentation.
Again, we note that the better results of depth completion methods with low-noise level is likely to be due to overfitting.

\subsection{Inference cost study} 

\begin{table}[ht]
  \centering
    \caption{Cost comparison for different architectures. \\
    (ZS), (FT) and (DC) means Zero-Shot, Fine-Tuned and Depth Completion, respectively.}
  \begin{tabular}{|c|c|c|}
  \hline
  Methods                                               & \#param (M) & runtime (ms) \\ \hline
  Metric3D V2 \cite{hu2024metric3d} (ZS)               & 412         & 194          \\
  Unidepth \cite{piccinelli2024unidepth} (ZS)          & 347         & 140          \\
  ZoeDepth \cite{bhat2023zoedepth} (FT)                & 335         & 113          \\
  Depth Anything \cite{yang2024depth} (FT)             & 335         & 113          \\
  Depth Anything V2 \cite{yang2024depthv2} (FT)         & 336         & 128          \\
  CompletionFormer \cite{zhang2023completionformer} (DC)&  84         & 150          \\
  BP-Net \cite{tang2024bilateral} (DC)                  &  90         & 103          \\ 
  Ours (Depth Anything \cite{yang2024depth} + rescaling) & 335         & 120          \\ \hline
  \end{tabular}%
  \label{tab:param_runtime}
  \end{table}

To complete our study, we analyze the inference cost of rescaling Depth Anything V1 affine-invariant disparity maps with the other zero-shot (ZS), fine-tuned (FT) monocular depth estimation or depth completion (DC) methods in \cref{tab:param_runtime}. 
For this purpose, we compare two informative and easy-to-access variables which are the number of parameters and the average inference runtime of these approaches for a single image. 
Note that the runtimes have been measured on the same NVIDIA GeForce RTX $3090$ GPU for each approach.
We notice that zero-shot methods are more costly at inference than our rescaling approach because they introduce extra modules such as a ConvGRU block in \cite{hu2024metric3d} or a camera module in \cite{piccinelli2024unidepth} that returns a dense representation of the camera calibration.
In contrast, fine-tuned methods have lower inference costs relative to ours thanks to their more complex training. 
As for depth completion methods, they are always lighter in terms of parameters which can be justified for architectures designed to be trained on a single dataset. 
Consequently, we could have expected a low inference time, which is what we observe with BP-Net but not with CompletionFormer \cite{zhang2023completionformer}.
It may be due to exotic modules in their architecture such as the Joint Convolutional Attention and Transformer block (JCAT). However, one may think that CompletionFormer could be sped up by running independent modules in parallel. 
As each image is processed independently, the computation overhead of our method can be even reduced when applied to a video stream. 
Indeed, since the neural network runs on the GPU and the RANSAC on the CPU, the inference of the neural network at step $t+1$ can overlap the execution of RANSAC at step $t$.

\subsection{Comparison with Depth Anything V2}

\begin{table}[ht]
  \centering
    \caption{Comparative study between Depth Anything V1 and Depth Anything V2 on the KITTI dataset \cite{kitti}.}
  \begin{tabular}{|c|c|cccc|}
  \hline
  \multirow{2}{*}{\begin{tabular}[c]{@{}c@{}}Rescaling \\ with\end{tabular}} & \multirow{2}{*}{\begin{tabular}[c]{@{}c@{}}Depth \\ Anything\end{tabular}} & \multicolumn{4}{c|}{KITTI \cite{kitti}} \\
                            &    & $\delta_1$ $\uparrow$ & AbsRel $\downarrow$ & RMS $\downarrow$ & $R^2$ $\uparrow$  \\ \hline
  \multirow{2}{*}{LiDAR 1}  & V1 & \textbf{0.891}        & \textbf{0.131}      & \textbf{3.096}    & \textbf{0.834} \\
                            & V2 & 0.733                 & 0.193               & 7.053             & 0.721 \\ \hline
  \multirow{2}{*}{LiDAR 16} & V1 & \textbf{0.967}        & \textbf{0.060}      & \textbf{2.695}    & \textbf{0.966} \\
                            & V2 & 0.961                 & 0.067               & 2.829             & 0.954 \\ \hline
  \multirow{2}{*}{LiDAR 32} & V1 & \textbf{0.967}        & \textbf{0.060}      & \textbf{2.673}    & \textbf{0.964} \\
                            & V2 & 0.962                 & 0.067               & 2.815             & 0.953 \\ \hline
  \end{tabular}%
  \label{tab:DAv1_vs_DAv2}
  \end{table}

\begin{figure}[ht]
    \includegraphics[width=0.48\textwidth]{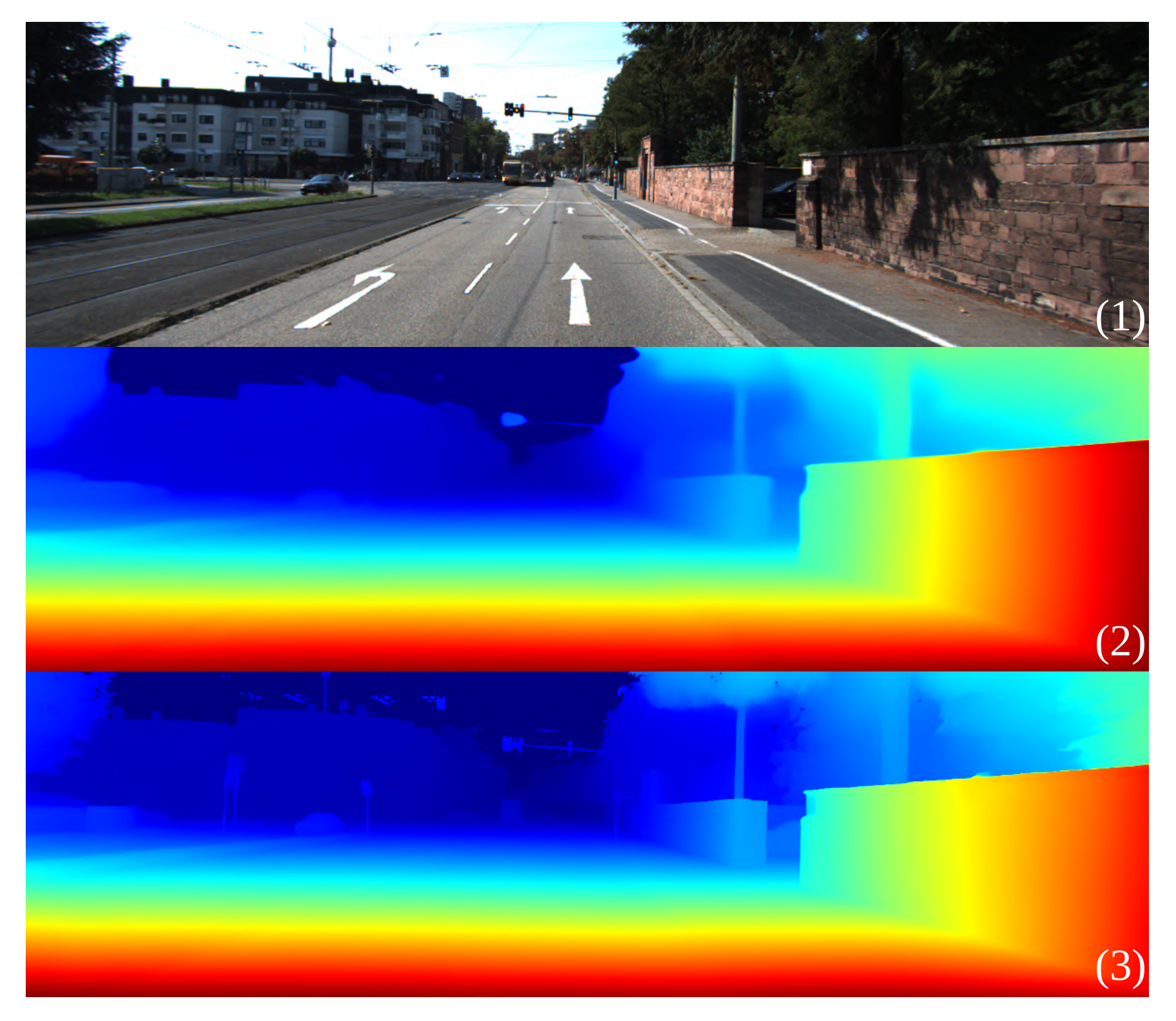}
    \caption{Qualitative comparison between Depth Anything V1 \cite{yang2024depth} and V2 \cite{yang2024depthv2}. From top to bottom: (1) image from the KITTI dataset \cite{kitti}, (2) the disparity map predicted by Depth Anything V1 and (3) the one of Depth Anything V2.}
    \label{fig:dav1_dav2}
\end{figure}

We compare performance between Depth Anything V1 \cite{yang2024depth} and Depth Anything V2 \cite{yang2024depthv2} on the KITTI dataset in \cref{tab:DAv1_vs_DAv2}.
We notice that Depth Anything V1 \cite{yang2024depth} slightly outperforms Depth Anything V2 \cite{yang2024depthv2} when used in our rescaling approach. 
Moreover, Depth Anything V1 has a higher coefficient of determination $R^2$ which corresponds to the proportion of the variance of the metric disparity that is explainable by the linear regression model parameterized with the $\alpha$ and $\beta$ that has been found. 
This means that while Depth Anything V2 manages to handle small details better than Depth Anything V1 (see \cref{fig:dav1_dav2}), the latter predicts more accurate disparity maps within an affine transformation.

\subsection{Ablation study}
  
\begin{table}[ht]
\centering
\caption{Ablation study to validate the need for estimating the rescaling parameters for each image}
\begin{tabular}{|c|ccc|}
\hline
\multirow{2}{*}{\begin{tabular}[c]{@{}c@{}}Rescaling\\ with\end{tabular}} & \multicolumn{3}{c|}{NYUv2 \cite{nyu}} \\
                 & $\delta_1$ $\uparrow$ & AbsRel $\downarrow$ & RMS $\downarrow$ \\ \hline
fixed parameters & 0.787                 & 0.142               & 1.240             \\
LiDAR 1 beam         & 0.939                 & 0.063               & 0.652             \\
LiDAR 16 beams        & \textbf{0.976}        & \textbf{0.039}      & \textbf{0.454}    \\
LiDAR 32 beams        & {\ul 0.974}           & {\ul 0.040}         & {\ul 0.461}       \\ \hline
                 & \multicolumn{3}{c|}{KITTI \cite{kitti}}                                      \\
                 & $\delta_1$ $\uparrow$ & AbsRel $\downarrow$ & RMS $\downarrow$ \\ \hline
fixed parameters & 0.908                 & 0.106               & 3.976             \\
LiDAR 1 beam         & 0.813                 & 0,152               & 6,032             \\
LiDAR 16 beams        & \textbf{0.966}        & \textbf{0.060}      & \textbf{2.755}    \\
LiDAR 32 beams         & \textbf{0.966}        & \textbf{0.060}      & {\ul 2.813}       \\
SFM (SIFT)       & 0.910                 & {\ul 0.103}               & 3.686             \\
SFM (OmniGlue)   & {\ul 0.921}                 & {\ul 0.103}               & 3.179             \\
\hline
\end{tabular}%
\label{tab:fixed_params}
\end{table}

\begin{table}[ht]
  \centering
    \caption{Ablation study to validate the use of RANSAC.}
  \resizebox{0.49\textwidth}{!}{%
  \begin{tabular}{|c|c|ccc|}
  \hline
  \multirow{2}{*}{Methods}                           & \multirow{2}{*}{RANSAC} & \multicolumn{3}{c|}{KITTI \cite{kitti}}                                      \\
                                                    &                         & $\delta_1$ $\uparrow$ & AbsRel $\downarrow$ & RMS $\downarrow$ \\ \hline
  \multirow{2}{*}{Ours w/ LiDAR 1 beam} & \xmark & \textbf{0.897} & \textbf{0.113} & 4.871 \\
                                                    & \cmark                  & 0.891                 & 0.131               & \textbf{3.096}    \\ \hline
  \multirow{2}{*}{Ours w/ LiDAR 16 beams} & \xmark                  & 0.966                 & 0.063               & 2.849             \\
                                                    & \cmark                  & \textbf{0.967}        & \textbf{0.06}       & \textbf{2.695}    \\ \hline
  \multirow{2}{*}{Ours w/ LiDAR 32 beams} & \xmark                  & 0.966                 & 0.063               & 2.852             \\
                                                    & \cmark                  & \textbf{0.967}        & \textbf{0.06}       & \textbf{2.673}    \\ \hline
  \multirow{2}{*}{Ours w/ SFM (SIFT)}     & \xmark                  & 0.001                 & 0.926               & 19.053            \\
                                                    & \cmark                  & \textbf{0.893}        & \textbf{0.103}      & \textbf{3.92}     \\ \hline
  \multirow{2}{*}{Ours w/ SFM (OmniGlue)} & \xmark                  & 0.859                 & 0.107               & 4.012             \\
                                                    & \cmark                  & \textbf{0.925}        & \textbf{0.093}      & \textbf{3.562}    \\ \hline
  \end{tabular}%
  }
  \label{tab:ablation}
 \end{table}
\cref{tab:fixed_params} compares dynamic rescaling, \emph{i.e.}, an estimation of the parameters $\alpha$ and $\beta$ for each image to a static rescaling with unique parameters for all the images of a dataset. In the latter case, the parameters are the mean $\alpha$ and $\beta$ obtained with the best rescaling (here with LiDAR 16 beams).
The results show that performing a rescaling for each image tends to provide better performance even if the camera does not change or if the image domain remains similar.

The ablation study in \cref{tab:ablation} aims to justify the use of a RANSAC algorithm \cite{fischler1981random} when estimating the parameters of the affine transformation to recover the metric depth. We observe that the RANSAC is beneficial most of the time, especially when the 3D reference points are not provided by a LiDAR. Indeed, structure-from-motion 
are more likely to generate outliers that would disturb a vanilla linear regression but would be filtered out by the RANSAC. 

\section{Conclusion}

In this paper, we provide a straightforward method to estimate monocular metric depth by rescaling Depth Anything V1 \cite{yang2024depth} affine-invariant disparity predictions using 3D reference points provided by an external sensor or technique.
By using Depth Anything V1 predictions whose weights are publicly available, we ensure generalization capacities to a large variety of image domains.
By leveraging RANSAC, our method is robust to noise in the sparse depth or in the disparity maps, allowing the use of low-quality sensors for metric depth estimation.
Thus, the solution we propose is adaptable to any camera calibration and does not need any fine-tuning of the monocular depth estimation neural network, which in practice also means that no costly creation of a dataset of the target domain is required.
For all these reasons, our approach is a good candidate for providing monocular metric depth at low cost.
To corroborate our claims we carry out experiments on standard depth estimation benchmarks that show that our approach is competitive with other zero-shot monocular depth estimation methods.
We also demonstrate superiority with respect to depth completion methods in downgraded mode.
Our future works will focus on confirming the advantages of our approach by comparing it with a high-resolution LiDAR in the context of off-road navigation with a real robot.

\balance
\bibliographystyle{IEEEtran}
\bibliography{IEEEabrv,biblio}

\end{document}